\documentclass[10pt, a4paper]{article}
\usepackage[utf8]{inputenc}
\usepackage[T1]{fontenc}
\usepackage{lmodern}
\usepackage[margin=2.5cm]{geometry}
\usepackage{amsmath, amssymb}
\usepackage{graphicx}
\usepackage{booktabs}
\usepackage[numbers,sort&compress]{natbib}
\usepackage[colorlinks=true,linkcolor=blue,citecolor=blue,urlcolor=blue]{hyperref}
\usepackage{cleveref}
\usepackage{xcolor}
\usepackage{subcaption}
\usepackage{multirow}
\usepackage{enumitem}
\usepackage{titlesec}
\usepackage{titling}
\usepackage{setspace}
\usepackage{caption}
\usepackage{float}
\pretitle{\centering\fontsize{14}{16}\selectfont\bfseries}
\posttitle{\par}

\titleformat{\section}
  {\normalfont\fontsize{14}{16}\bfseries}
  {\thesection}{1em}{}

\titleformat{\subsection}
  {\normalfont\fontsize{11}{13}\bfseries}
  {\thesubsection}{1em}{}

\title{\textbf{A Multi-Agent Framework for Automated Coarse-Grained Molecular Dynamics of Polymers}}

\author{
  Joohee Choi\textsuperscript{1†},
  Junhyeong Lee\textsuperscript{2†},
  Seunghwa Ryu\textsuperscript{1,2,3,*}
  \\[0.5em]
  \textsuperscript{1}Department of Mechanical Engineering,\\
  Korea Advanced Institute of Science and Technology (KAIST),\\
  Daejeon 34141, Republic of Korea
  \\[0.5em]
  \textsuperscript{2}KAIST InnoCORE PRISM-AI Center,\\
  Korea Advanced Institute of Science and Technology (KAIST),\\
  Daejeon 34141, Republic of Korea
  \\[0.5em]
  \textsuperscript{3}Department of AX,\\
  Korea Advanced Institute of Science and Technology (KAIST),\\
  Daejeon 34141, Republic of Korea
  \\[0.5em]
}
\date{}

\begin{document}
\maketitle

\noindent
\textsuperscript{†}These authors contributed equally.\\
* Corresponding author: ryush@kaist.ac.kr

\bigskip

% ─── Abstract ────────────────────────────────────────────────────────────────
\begin{abstract}
Coarse-grained (CG) molecular dynamics extends polymer simulation beyond the scales accessible to all-atom (AA) methods, but bottom-up CG modeling is laborious. The CG resolution is a design choice, so a transferable parameter set is generally not available and the potentials are derived anew for each polymer mapping. Here we present CGMas, a multi-agent framework that automates topology construction, equilibration, mapping, potential derivation, and validation from a natural-language specification of the polymer and target resolution. A large-language-model (LLM) reasoning agent infers the AA topology from polymer name, while layered self-correction resolves physical errors common to unsaturated, heteroatom-containing, and polar polymers. Downstream agents equilibrate the system, map it onto CG representation, derive potentials through Boltzmann inversion, and benchmark the model against its atomistic reference. CGMas completed all 27 homopolymer and copolymer tasks, matched the AA density to within 5\% in 22, and reduced simulation from 38-88 min to 1 min, establishing agentic LLMs as a route to automated polymer coarse-graining.
\end{abstract}

% ─── 1. Introduction ─────────────────────────────────────────────────────────
\doublespacing
\section{Introduction}
\label{sec:intro}

Predicting the behavior of a polymer from its chemistry is a central objective of computational materials design~\citep{gartner2019modeling, chen2021polymer}. Molecular dynamics (MD) simulations is the standard route to such structure-property relationships as glass transition, crystallization, chain diffusion, and plastic deformation~\citep{kremer1990dynamics, han1994glass}. All-atom (AA) force-fields such as OPLS-AA, CHARMM, and PCFF describe molecular interactions with high fidelity at a high computational cost~\citep{jorgensen1996opls, mackerell1998all, sun1995ab}. AA simulations of polymer systems are typically limited to a few hundred nanoseconds of order $10^{5}$ atoms, which corresponds to a simulation box about $10$~nm on a side~\citep{peter2009multiscale}.

Coarse-grained (CG) models were developed to close this gap~\citep{mullerplathe2002coarse, choi2026coarse}. In a CG model, a group of atoms is replaced by a single effective interaction site, or bead~\citep{tschop1998simulation}. Removing the atomistic degrees of freedom reduces the number of pair interactions and smooths the  energy landscape, allowing a larger integration time step~\citep{marrink2007martini}. The softer interactions also reduce the effective friction, so conformational relaxation proceeds faster than in the AA simulation~\citep{harmandaris2006hierarchical}. These effects extend the accessible time and length scales by one to two orders of magnitude at fixed computational cost~\citep{peter2009multiscale}.

The CG resolution is not fixed, but is a design choice~\citep{milano2005mapping}. The user decides how finely the polymer is represented and the atoms are grouped into beads according to a mapping that trades spatial detail for computational efficiency., with the interaction between bead beads specified~\citep{fritz2009coarse}. Bottom-up CG force-fields are obtained directly from AA reference data, so that the CG model reproduces the structural and thermodynamic properties of its atomistic reference~\citep{reith2003deriving, izvekov2005effective, shell2008relative}. 

The freedom is also the source of the central difficulty. Because the mapping is user-defined, a universal parameter set is not available, and potentials need to be derived for every polymer and every mapping~\citep{louis2002beware}. Doing so requires a long and interdependent sequence of steps, from building a chemically correct AA topology and equilibrating an amorphous system, through choosing a bead mapping and accumulating the bonded and non-bonded distributions, to inverting them into potentials and validating the assembled model~\citep{harmandaris2006hierarchical, milano2005mapping}. Each step requires domain-specific knowledge. Errors in atom typing, partial charges, or equilibration readily arise, often remain undetected, and propagate through subsequent stages, resulting in a physically incorrect model.

Existing software tools partially address this challenge~\citep{jewett2021moltemplate, fortunato2017pysimm}. VOTCA and PyCGTOOL take the atomistic input and the bead mapping as given, so the stages where errors originate remain with the user~\citep{ruhle2009versatile, graham2017pycgtool}. \texttt{martinize2} constructs CG topologies only from the predefined building blocks of a biomolecular force-field, which do not cover synthetic polymers~\citep{kroon2025martinize2, souza2021martini}. Synthetic polymers lack a comparable inventory, as chemistry varies with tacticity, sequence, and branching, while material varies with molecular weight, morphology, and processing conditions~\citep{fritz2009coarse, audus2017polymer, choi2022predicting}.

We hypothesize that the bottleneck lies not in the individual computational steps, but in the mismatch between fragmented, tool-specific automation and the integrative, multi-step reasoning that an expert applies across the pipeline as a whole. Large language models (LLMs) can supply that reasoning~\citep{jablonka2023fourteen}. Recent work shows that they can reason about the chemical structure, generate molecular representations, and draft simulation inputs~\citep{kuenneth2023polybert, kang2024chatmof, shi2025finetuned}. However, their fluency is not evidence of physical validity, so the reasoning needs to be paired with an explicit check at each step~\citep{ramos2025review}.

Agentic systems meet both requirements by coordinating specialized agents into a single workflow, so that reasoning and verification are assigned to different components~\citep{ramos2025review}. This approach has begun to reach adjacent domains of computational chemistry and materials science, including tool-augmented chemistry agents and autonomous experimentation, workflow automation in molecular dynamics and quantum chemistry, and multi-agent systems for materials design and scientific discovery~\citep{bran2024chemcrow, boiko2023coscientist, campbell2026mdcrow, shi2025finetuned, zou2025elagente, ghafarollahi2025atomagents, ghafarollahi2025sciagents}. But, none of these frameworks addresses the bottom-up coarse-graining of polymers, where a chemically valid atomistic topology is first generated and then transformed into a mapping-specific CG force field, requiring physical validation required.

Here we introduce CGMas (Coarse-Graining Multi-agent system), an autonomous framework that executes the entire AA-to-CG pipeline from natural-language input. The user supplies a polymer name, with a plain-language description of the desired bead resolution. The CGMas couples LLM-based chemical reasoning to a suite of purpose-built, physics-aware tools, each responsible for one stage of the workflow.

Its central contribution is that the CG force field itself is derived autonomously for whatever resolution the user asks for. From a plain-language description of the desired beads, CGMas defines the mapping, enumerates the interaction types it implies, accumulates the corresponding distributions from the atomistic trajectory, and inverts and fits them into a screened set of potentials. This is possible without a user-supplied topology because a reasoning agent first generates the OPLS-AA~\citep{jorgensen1996opls, dodda2017ligpargen} constitutional repeat unit from the polymer name, assigning each atom the force-field type and partial charge appropriate to its local structure and specifying the connectivity, with a layered self-correction loop that diagnoses and repairs the failures common to unsaturated, aromatic, polar, and heteroatom-containing repeat units.

Its central contribution is the autonomous generation of a chemically valid OPLS-AA topology. From the polymer name, a reasoning agent identifies the constitutional repeat unit, assigns each atom the force-field type, and the partial charge appropriate to its local structure, and specifies the connectivity. In particular, for repeat units with unsaturated bonds, aromatic and polar chains, and heteroatoms, a layered self-correction loop diagnoses each failure and repairs the topology, encoding generalizable physical reasoning. Downstream agents then carry out the remaining stages autonomously, from unit cell construction and equilibration through the derivation of CG potentials to validation against the atomistic reference from which the model is derived. The framework supports homopolymers and copolymers across random, alternating, block, and periodic architectures.

We evaluated CGMas on a benchmark of 27 polymer tasks spanning five levels of chemical difficulty and show that it reproduces the equilibrium density of its atomistic references without requiring user expertise in force-field parameterization or simulation scripting. By condensing an expert workflow into a conversational interface, CGMas lowers the barrier to bottom-up CG modeling of polymers and suggests a general strategy for physics-based, AI-driven automation of molecular-simulation pipelines.

% ─── 2. Results ───────────────────────────────────────────────────────────
\section{Results}
\label{sec:results}

\subsection{CGMas framework}
\label{subsec:framework}

\begin{figure}[H]
  \centering
  \includegraphics[width=\linewidth]{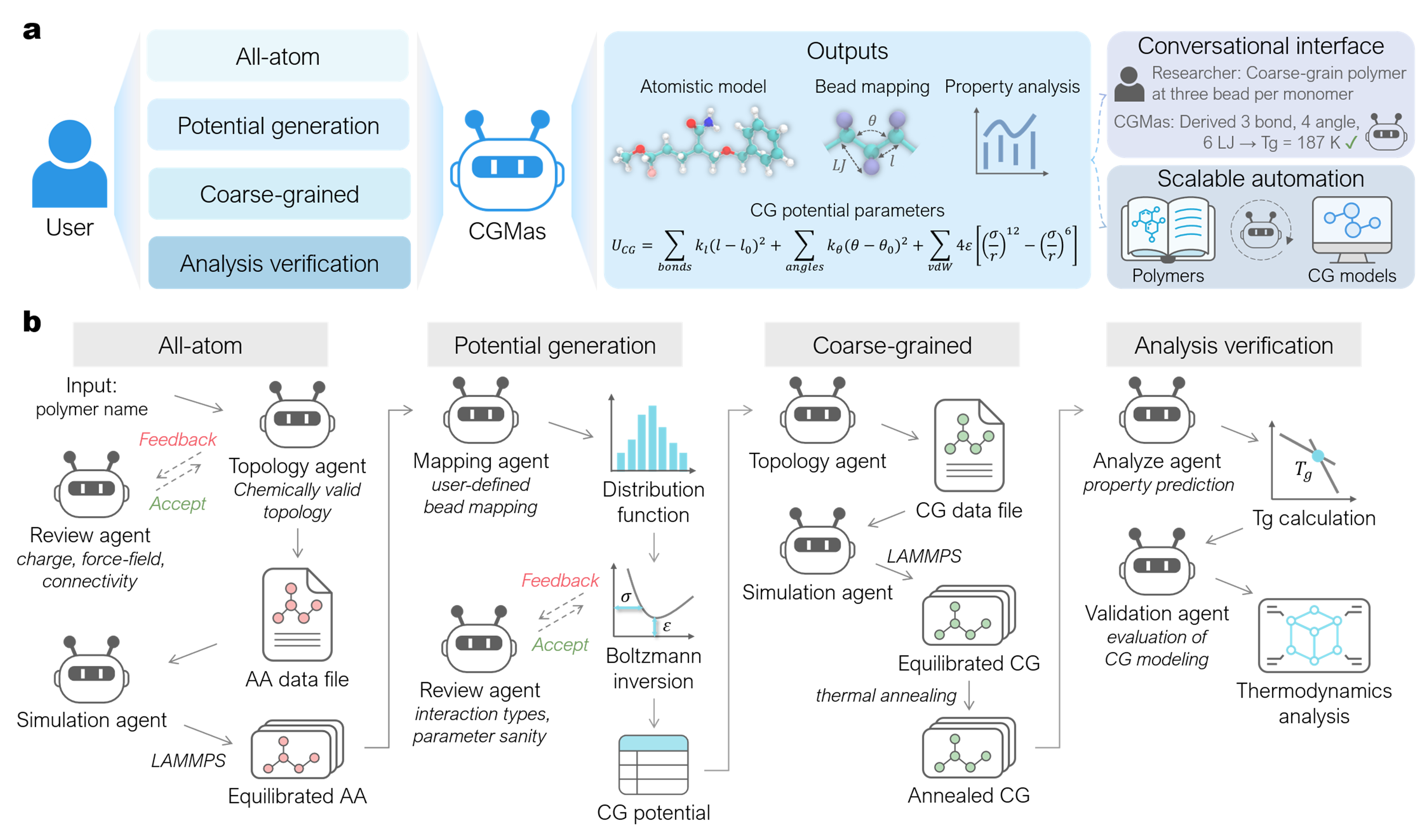}
  \caption{\textbf{a} Illustration of using CGMas to automate polymer coarse-graining. Users interact with CGMas through a conversational interface across four tasks, including all-atom, potential generation, coarse-grained, and analysis verification. Given a polymer name and target resolution, CGMas generates an atomistic model, bead mapping, CG potential parameters, and thermodynamic properties, enabling automated CG model construction across diverse polymer chemistries. \textbf{b} Overview of CGMas, a multi-agent framework in which reasoning agents for topology generation, review, mapping, simulation, analysis, and validation perform chemical inference and quality control, while deterministic tools execute molecular simulations, distribution analysis, Boltzmann inversion, and property calculations. Generator–reviewer feedback loops correct errors during topology construction and potential generation prior to validation of the CG model and its predicted properties.}
  \label{fig:workflow}
\end{figure}

We designed CGMas to cast the bottom-up coarse-graining pipeline as a sequence of specialized agents, each owning one stage and communicating through a shared, persistent description of the simulation state. Figure~\ref{fig:workflow}a illustrates this organization. A state-machine controller routes information between the agents, bounds the correction loops, and pauses for user confirmation at the scientifically critical checkpoints, with the orchestration details given in \Cref{sec:methods}. We divide the labor between two kinds of components. Reasoning agents, backed by an LLM, perform the open-ended chemical inference that an expert would otherwise supply, whereas deterministic tools, implemented as purpose-built modules, perform the exact and reproducible computation. Importantly, we wrote these tools for the present work rather than assembling them from existing libraries, so that all stages share a common data representation and expose the diagnostics that the reasoning agents require for self-correction. This separation allows the LLM to contribute flexible judgment where it is needed while keeping every numerical operation transparent and physically constrained.

The pipeline proceeds through six stages, which Fig.~\ref{fig:workflow}b groups into all-atom, potential generation, coarse-grained, and analysis and verification modules.

\begin{itemize}
  \item Input: a reasoning agent parses the polymer name, detects whether the system is a homopolymer or a copolymer, and, for a copolymer, resolves its architecture.
  \item All-atom topology: a reasoning agent generates the OPLS-AA constitutional repeat unit, which a layered self-correction loop and an independent review agent validate and repair (\Cref{subsec:methods_topology}). Notably, this is the stage that carries the most chemical inference and therefore the most extensive correction machinery.
  \item All-atom simulation: a deterministic tool packs an amorphous multi-chain system, which is then equilibrated under the confirmed system, energy minimization, and anneal protocol.
  \item Mapping and potential derivation: a reasoning agent translates the natural-language resolution description into a bead definition, a deterministic tool computes the bead trajectories and inverts the accumulated distributions into potentials, and a review agent screens the fitted parameters for unphysical values.
  \item Coarse-grained simulation: a deterministic tool assembles the CG model from the fitted parameters and equilibrates it under the same thermal protocol as the all-atom system.
  \item Analysis and validation: a deterministic tool extracts the glass-transition temperature from a cooling scan, and a validation agent compares the CG density and $T_g$ with the atomistic reference produced in the same session.
\end{itemize}

We benchmark the CG model against the atomistic reference generated within the same session, which residual deviation reflects the coarse-graining step. As depicted in Fig.~\ref{fig:workflow}b, five checkpoints allow an expert to intervene where scientific judgment matters, namely polymer identity, box and protocol parameters, bead mapping, potential approval, and final validation. All computation between checkpoints proceeds without manual effort. Because we persist the full state, a session can be paused at any checkpoint and resumed without re-running the upstream stages, and we retain a log of every agent decision for reproducibility. Taken together, these design choices place the expert judgment at the points where it is most consequential and leave the remainder of the pipeline to run unattended.

\subsection{Human-AI collaboration results}
\label{subsec:example}

\begin{figure}[H]
  \centering
  \includegraphics[width=\linewidth]{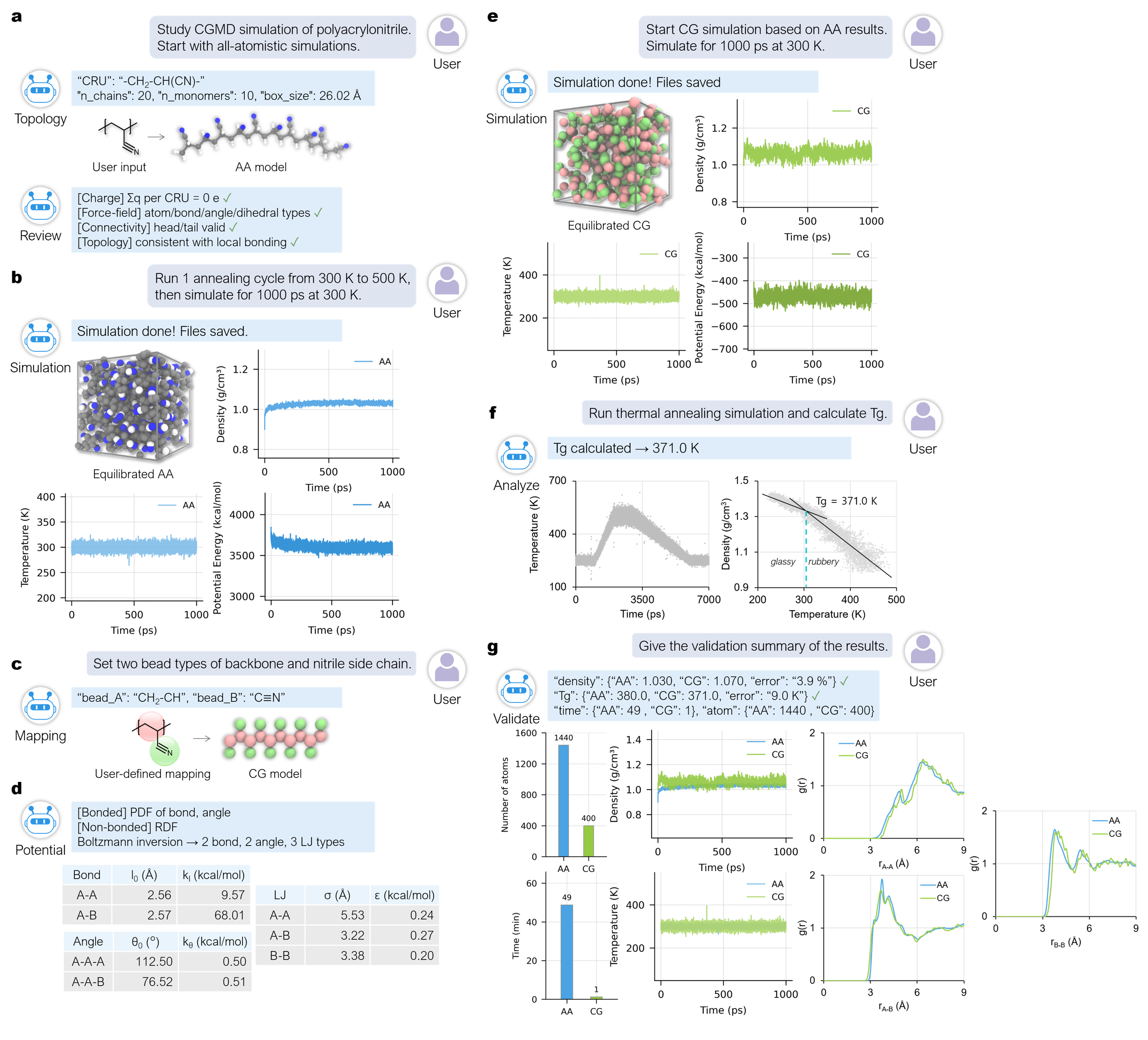}
  \caption{Representative interactive process of CGMas for polyacrylonitrile at two beads per monomer. \textbf{a} From a single polymer-name query, CGMas autonomously performs topology generation and review, \textbf{b} all-atom bulk equilibration, \textbf{c} natural-language bead mapping, \textbf{d} Boltzmann-inversion potential derivation, \textbf{e} CG equilibration, \textbf{f} $T_g$ extraction from a thermal-annealing run, and \textbf{g} validation against the AA reference.}
  \label{fig:interaction}
\end{figure}

To make the workflow concrete, we trace a complete interactive CGMas session for polyacrylonitrile (PAN) at a resolution of two beads per monomer, which Fig.~\ref{fig:interaction} illustrates. The user opens with ``Study CGMD simulation of polyacrylonitrile. Start with all-atomistic simulations.'' The topology agent identifies PAN as a vinyl homopolymer with the CRU --CH$_2$--CH(CN)-- and proposes a default system of 20 chains $\times$ 10 repeat units, comprising 1{,}440 atoms including the chain-terminating hydrogens, in a cubic box of side 26.02~\AA. The review agent then verifies the generated topology against four independent criteria, which are charge neutrality of the CRU ($\sum q = 0\,e$), completeness of the atom, bond, angle, and dihedral force-field types, validity of the head and tail connection atoms, and consistency of the atom-type assignment with the local bonding environment, and approves it. Importantly, each criterion returns the specific discrepancy rather than a generic failure flag, which we expect to be the reason the topology stage runs unattended for this chemistry (Fig.~\ref{fig:interaction}a).

On the instruction ``Run 1 annealing cycle from 300~K to 500~K, then simulate for 1000~ps at 300~K,'' the simulation agent equilibrates the AA system in LAMMPS under the thermal annealing protocol. We assessed equilibration based on the convergence of the thermodynamic properties. Specifically, the density regulated by the barostat, the temperature controlled by the thermostat, and the total potential energy all converged to steady mean values and subsequently showed stationary fluctuations, with the density plateauing at 1.03~g\,cm$^{-3}$ (Fig.~\ref{fig:interaction}b). These evaluations confirm that the AA system is fully equilibrated before any mapping is performed, which matters because the accumulated distributions inherit any residual drift of the reference trajectory.

\begin{figure}[H]
  \centering
  \includegraphics[width=\linewidth]{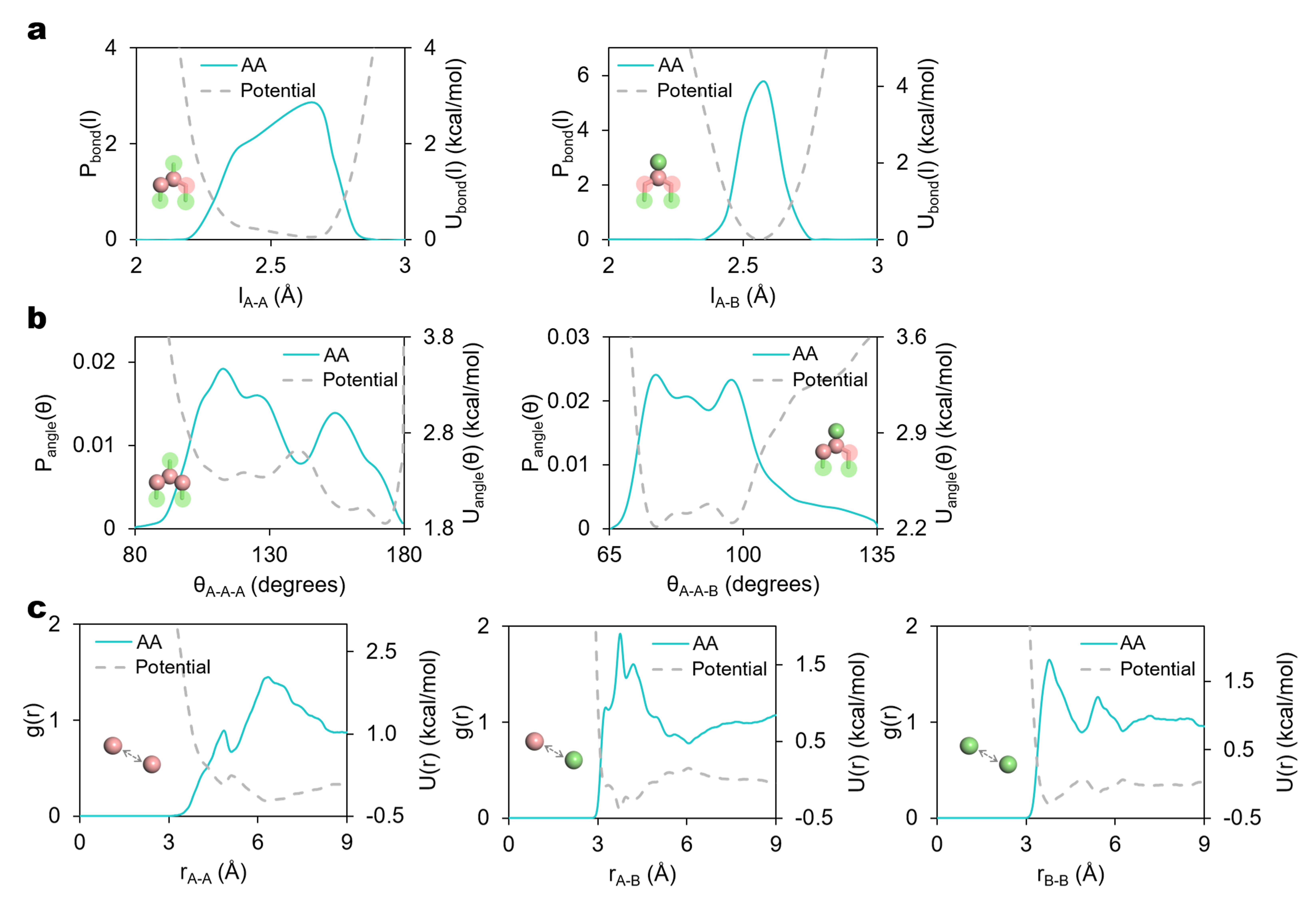}
  \caption{Probability distribution functions of the \textbf{a} bond lengths and \textbf{b} bending angles obtained from PAN of AA models at 300 K, and the corresponding potential energy curve. \textbf{c} Radial distribution functions obtained from PAN of AA models at 300 K.}
  \label{fig:potentials}
\end{figure}

The user then specifies the CG resolution in plain language, ``Set two bead types of backbone and nitrile side chain,'' which the mapping agent translates into bead~A for the CH$_2$--CH backbone and bead~B for the C$\equiv$N side group, yielding a 400-bead CG representation (Fig.~\ref{fig:interaction}c). From the mapped trajectory, the potential stage accumulates the bonded probability distributions and the non-bonded radial distribution functions (RDF), and Boltzmann inversion returns two bond types, two angle types, and three Lennard-Jones (LJ) pair types. Representative parameters are $l_0 = 2.56$~\AA\ and $k_l = 9.57$~kcal\,mol$^{-1}$\,\AA$^{-2}$ for the A--A backbone bond, $\theta_0 = 112.50^\circ$ for the A--A--A backbone angle, and $\sigma = 5.53$, 3.22, and 3.38~\AA\ with $\varepsilon = 0.24$, 0.27, and 0.20~kcal\,mol$^{-1}$ for the A--A, A--B, and B--B pairs, respectively. The review agent confirms that all expected interaction types are present and that their parameters lie in a physically reasonable range (Fig.~\ref{fig:interaction}d). Figure~\ref{fig:potentials} examines this potential-derivation stage in detail, overlaying each sampled distribution with corresponding CG potentials for the bond length, bending angle, and non-bonded types, respectively.

We then equilibrate the CG model for 1000~ps at 300~K under the same protocol. As for the AA system, the density, the temperature, and the total potential energy converge to steady mean values and fluctuate stationarily thereafter, with the density plateauing at 1.07~g\,cm$^{-3}$, which indicates that the CG system also reaches thermodynamic equilibrium. Notably, the mapped structure is preserved as well as the density (Fig.~\ref{fig:interaction}e). As shown in Fig.~\ref{fig:interaction}g, the CG pair distribution functions overlay those of the AA reference for all three bead pairs, reproducing both the excluded-volume onset and the position and height of the first coordination shell, which suggests that the fitted LJ parameters retain the local packing of the atomistic model and not only its average density. A subsequent thermal annealing run, analyzed through a bilinear fit of the cooling branch of the density-temperature curve, locates the rubbery-to-glassy transition at $T_g^{\mathrm{CG}} = 371.0$~K, compared with $T_g^{\mathrm{AA}} = 380.0$~K obtained under the identical protocol and chain length (Fig.~\ref{fig:interaction}f).

The validation agent summarizes the session as follows (Fig.~\ref{fig:interaction}g). The CG density deviates from the AA reference by 3.9\% (1.070 against 1.030~g\,cm$^{-3}$), which passes the tolerance of $\delta_\rho = 5\%$, and the two glass-transition temperatures agree to within 9.0~K. The slightly lower CG value is consistent with the reduced conformational friction of the mapped model, which may allow relaxation to persist to lower temperatures at the same cooling rate. As shown in Fig.~\ref{fig:interaction}g, the AA model required 49~min to complete the equilibration run, whereas the corresponding CG model completed the same protocol within 1~min, an approximately fifty-fold reduction obtained by reducing the system from 1{,}440 atoms to 400 beads. Also, the molecular structural characteristics were quantified by RDF. The RDFs of the equilibrated CG simulation overlay those of the mapped AA trajectory, indicating that the CG model preserves the essential structural characteristics of the atomistic system while providing a substantial reduction in computational cost. These results demonstrate that the CG model reproduces the equilibrium density, the glass-transition temperature, and the pair correlations of its atomistic reference while following the same equilibration process at a substantially lower computational cost, and that a single natural-language session delivers both the CG model and the reference against which it is judged.

\subsection{Tasks for evaluating the CGMas framework}
\label{subsec:task}

\begin{figure}[H]
  \centering
  \includegraphics[width=0.9\linewidth]{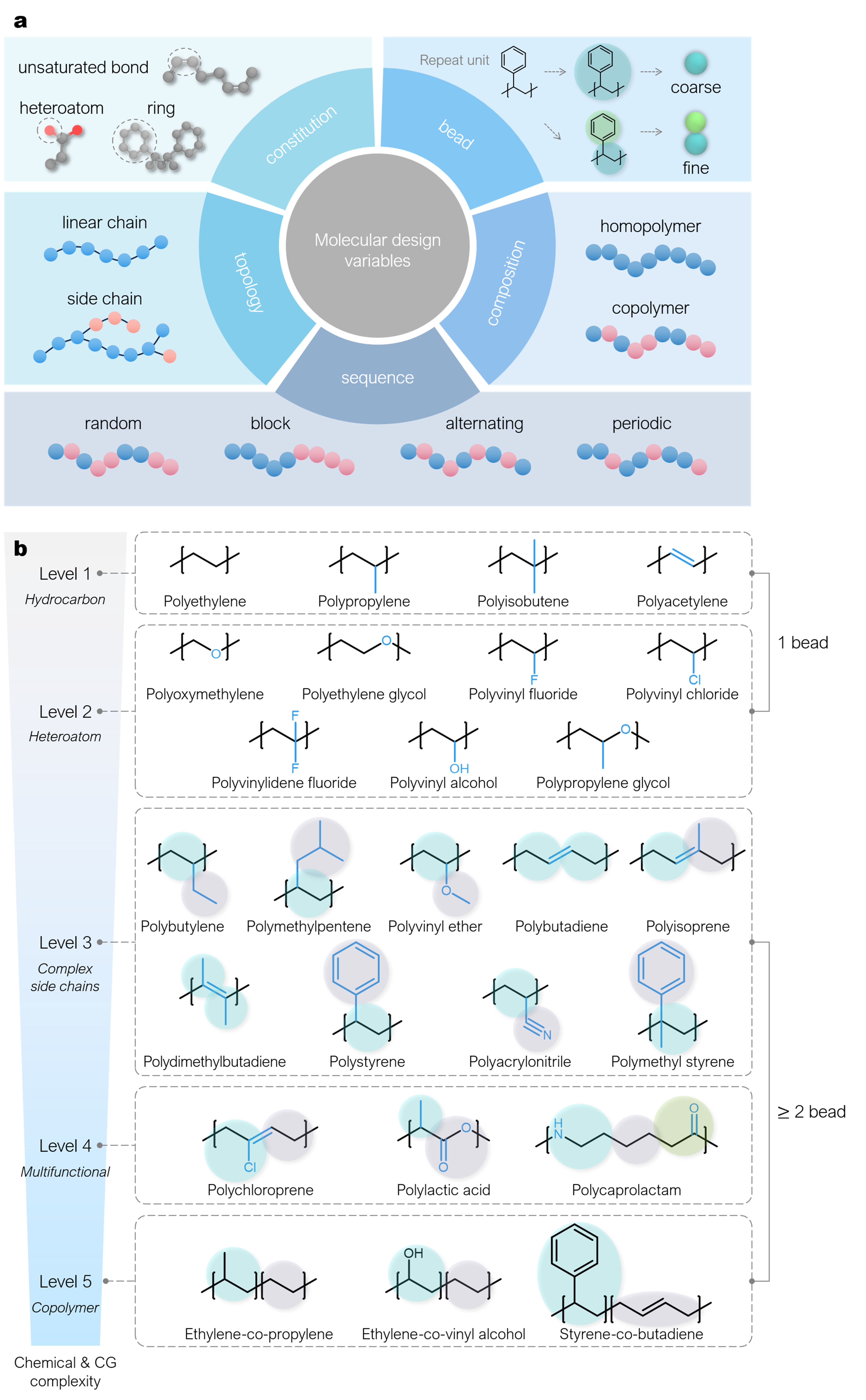}
  \caption{\textbf{a} Molecular modeling and design variables considered in polymer simulations, defiend by five categories. \textbf{b} The 27 benchmark tasks arranged by increasing chemical and coarse-graining complexity: hydrocarbons (Level 1), heteroatom-substituted repeat units (Level 2), complex side chains (Level 3), multifunctional repeat units (Level 4), and copolymers (Level 5). Levels 1-2 use one bead per monomer, and Levels 3-5 require two or more bead types.}
  \label{fig:tiers}
\end{figure}

Polymer space is not a list of molecules but a product of several largely independent design variables. Figure~\ref{fig:tiers}a summarizes the variables around which we organized the benchmark to generalize our pipeline. The constitution of the repeat unit, which includes unsaturation, heteroatoms, and rings, fixes which atom types and partial charges have to be assigned and therefore loads on topology generation. Whether the repeat unit carries a side chain, and how bulky that side chain is, fixes how the unit is partitioned into beads, since a side group large enough to warrant its own interaction site raises the bead count and with it the number of bonded and non-bonded types to be derived. The composition and the comonomer sequence fix whether one repeat unit or several are involved and in what order they are appended, and therefore load on the independent treatment and subsequent merging of the components. Because these variables act on different stages of the pipeline, the benchmark samples each of them rather than accumulating chemistries along a single axis.

We then ordered the resulting 27 tasks into a single five-level ramp (Fig.~\ref{fig:tiers}b), so that the pass rate can be read as a function of chemical complexity rather than as one aggregate number. Automated coarse-graining is demanding for two largely independent reasons, since the AA topology has to be generated correctly from the chemical structure and the CG representation has to be mapped and parameterized faithfully. The ramp stresses both. Levels 1 and 2 comprise 11 tasks that are mapped at one bead per monomer and primarily tax topology generation, whereas Levels 3 to 5 comprise 16 tasks that require two or more bead types and additionally tax bead mapping and potential derivation.

The five levels increase in chemical complexity as follows. Level 1 contains four hydrocarbons (polyethylene, polypropylene, polyisobutylene, and polyacetylene), of which the first three add methyl side chains to an otherwise linear carbon backbone, changing the substitution state of the backbone carbon from secondary to tertiary and then quaternary, while polyacetylene additionally requires sp$^2$ carbon types and matching hydrogens for its conjugated backbone. Level 2 introduces oxygen and halogen substituents in seven acyclic repeat units (polyoxymethylene, poly(ethylene oxide), poly(vinyl fluoride), poly(vinyl chloride), poly(vinylidene fluoride), poly(vinyl alcohol), and poly(propylene oxide)), in which chemically distinct heteroatom environments are easily interchanged and a single wrong assignment breaks charge neutrality.

Level 3 raises the complexity of the pendant group itself. Where the side chain of Level 1 was a single methyl, here it becomes a chemical unit in its own right, growing to an ethyl group in poly(1-butene) and an isobutyl group in poly(4-methyl-1-pentene), whose internal carbons differ in substitution state, acquiring a heteroatom in the methoxy group of poly(methyl vinyl ether), and closing into a ring in the phenyl group of polystyrene and poly($\alpha$-methylstyrene). The ring is the qualitative step, since a closed cycle cannot be built by sequential torsion sampling and is instead identified before placement and laid out as a regular planar polygon. The level also moves unsaturation out of the substituent and into the main chain, so that poly(1,3-butadiene), polyisoprene, and poly(2,3-dimethyl-1,3-butadiene) alternate sp$^2$ and sp$^3$ carbons along the backbone, with a methyl bonded directly to an sp$^2$ carbon in the latter two, and polyacrylonitrile adds sp hybridization through its nitrile carbon. Because each of these groups is large enough to warrant its own interaction site, the typing decisions become coupled to the mapping and the number of interaction types to be derived increases accordingly.

Level 4 moves the heteroatoms from the substituent into the backbone, so that the main chain is no longer a uniform carbon skeleton but changes chemistry along its length. Polychloroprene marks the transition, combining the sp$^2$ backbone of Level 3 with a chlorine whose environment differs from the sp$^3$ halogens of Level 2. Poly(lactic acid) then introduces an ester linkage, in which the carbonyl carbon, the doubly bonded oxygen, and the ester oxygen occupy three distinct environments whose charges balance against one another within a single repeat unit. Polycaprolactam adds an amide linkage, placing a hydrogen-bearing nitrogen next to the carbonyl, and its five-carbon methylene segment makes the repeat unit long enough to be partitioned into three beads. Because the bead boundaries now cut through a backbone whose chemistry varies along it, the typing and the mapping are coupled more tightly than for the vinyl polymers of the lower levels.

Level 5 extends to four copolymers that differ in comonomer sequence rather than in chemistry. These are poly(styrene-co-butadiene) with a triblock sequence of three styrene, four butadiene, and three styrene units, poly(ethylene-co-propylene) with a random sequence, and poly(ethylene-co-vinyl alcohol) in an alternating sequence of five ethylene and vinyl alcohol pairs and in a periodic sequence of three ethylene-ethylene-vinyl alcohol triads. In each copolymer, both comonomers are defined and self-corrected independently before the components are merged into a consistent model.

\subsection{Autonomous coarse-grained model generation}
\label{subsec:auto}

\begin{figure}[H]
  \centering
  \includegraphics[width=\linewidth]{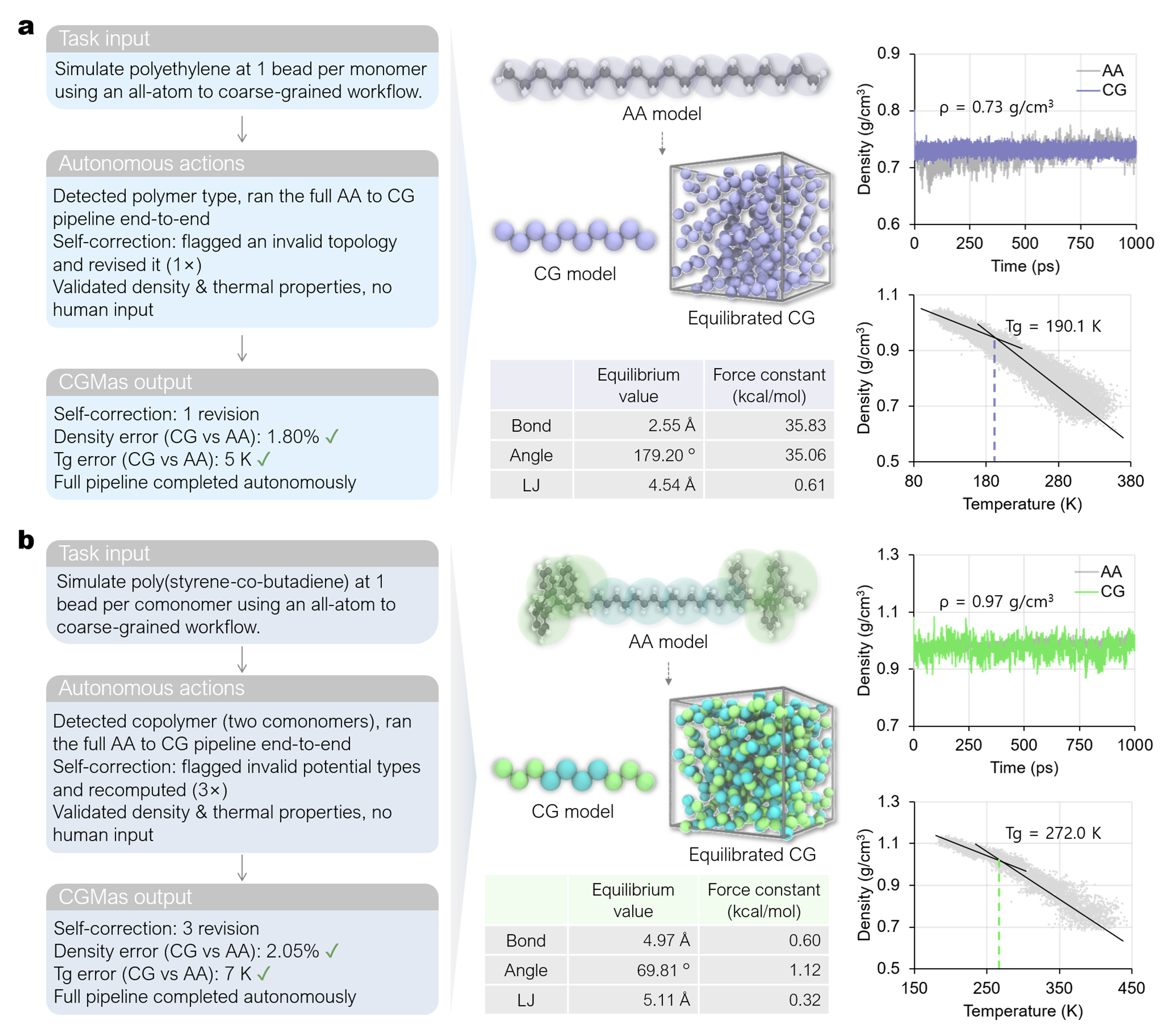}
  \caption{CGMas's capability to autonomously conduct coarse-graining tasks from a single query, without human intervention. \textbf{a} Polyethylene at one bead per monomer. \textbf{b} Poly(styrene-co-butadiene) at one bead per comonomer.}
  \label{fig:autonomous}
\end{figure}

The human-confirmation checkpoints in CGMas are optional safeguards rather than mandatory steps, so that a session in which the default parameters and the default bead mapping are accepted runs from a single query to a validated CG model with no further intervention. Figure~\ref{fig:autonomous} illustrates two such unattended runs, which we chose to bracket the difficulty range of the benchmark.

We first issued the query ``Simulate polyethylene at 1 bead per monomer using an all-atom to coarse-grained workflow'' (Fig.~\ref{fig:autonomous}a). CGMas detected the polymer type, ran the full AA-to-CG pipeline end to end, flagged one invalid topology during self-correction and revised it, and validated the resulting model without human input. The CG model, with $r_0 = 2.55$~\AA, $\theta_0 = 179.2^\circ$, and $\sigma = 4.54$~\AA, equilibrates to a density of 0.73~g\,cm$^{-3}$, within 1.80\% of its AA reference, and yields $T_g^{\mathrm{CG}} = 190.1$~K within 8~K of the AA value obtained under the identical protocol.

We then submitted a query targeting poly(styrene-co-butadiene) at one bead per comonomer (Fig.~\ref{fig:autonomous}b). CGMas detected the two-comonomer architecture, generated and corrected both repeat units independently, and completed the same pipeline autonomously. Notably, the three revisions in this run were triggered by invalid potential types rather than by an invalid topology, which indicates that the failure mode shifts from topology generation to potential derivation once two bead types are present. The resulting model equilibrates to a density of 0.97~g\,cm$^{-3}$, within 2.05\% of its AA reference, and yields $T_g^{\mathrm{CG}} = 272.0$~K within 12~K. Taken together, the two runs demonstrate that unattended operation extends from the simplest homopolymer to a copolymer requiring multi-component topology generation and merging.

These results demonstrate that the CG model reproduces the equilibrium density, the glass-transition temperature, and the pair correlations of its atomistic reference while following the same equilibration process at a substantially lower computational cost, and that a single natural-language session delivers both the CG model and the reference against which it is judged. In effect, the expert-dependent work of deriving a faithful CG model is performed automatically, after which the model provides the scale and speed that motivated coarse-graining in the first place.

\subsection{Performance of CGMas across difficulty levels}
\label{subsec:perform}

\begin{figure}[H]
  \centering
  \includegraphics[width=\linewidth]{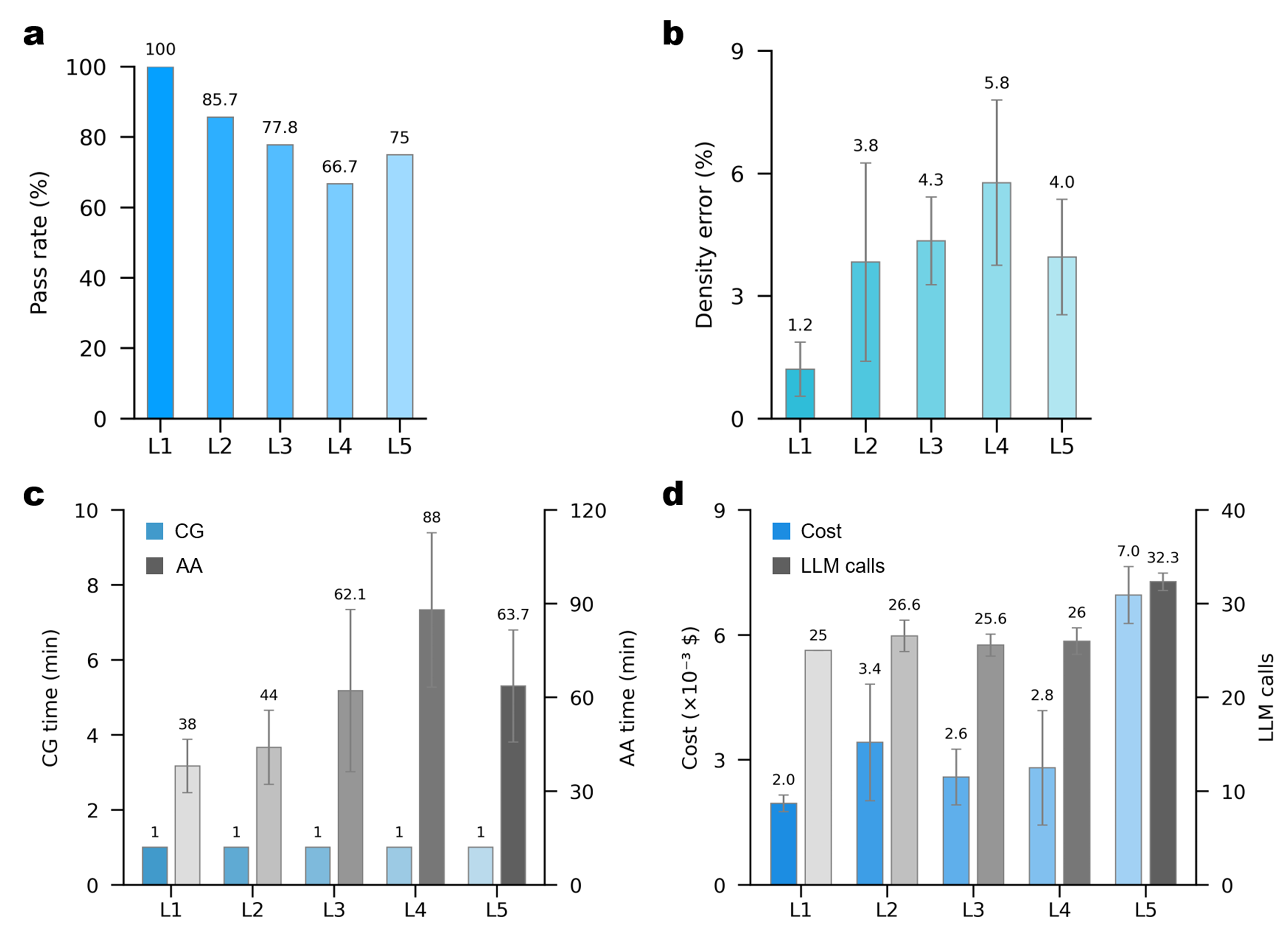}
  \caption{Benchmark performance of CGMas across the five difficulty levels. \textbf{a} Pass rate, defined as the fraction of tasks whose CG density agrees within 5\% of the AA reference. \textbf{b} Density error, defined as the mean absolute deviation of the CG density from the AA reference. \textbf{c} Total simulation time of the CG models compared with the AA reference. \textbf{d} LLM cost and number of LLM calls per task.}
  \label{fig:benchmark}
\end{figure}

Figure~\ref{fig:benchmark} summarizes CGMas performance across the five difficulty levels. We define the end-to-end pass rate as the fraction of tasks whose CG equilibrium density falls within 5\% of the all-atom reference. All 27 tasks ran to completion, so the tasks that did not pass exceeded this band rather than terminating before a CG model was produced. As shown in Fig.~\ref{fig:benchmark}a, the pass rate is 100\% for Level 1 and decreases with chemical complexity to 85.7\% for Level 2, 77.8\% for Level 3, and 66.7\% for Level 4, before recovering to 75.0\% for the Level 5 copolymers, giving an overall pass rate of 22 of 27 tasks (81.5\%). The monotonic decline across Levels 1 to 4 indicates that repeat-unit chemistry, rather than system size or simulation protocol, is the dominant driver of difficulty. Notably, the recovery at Level 5 suggests that copolymer merging is handled robustly once each comonomer topology is individually valid, so that the difficulty of a copolymer is set by its constituent chemistries rather than by the merging itself.

The density accuracy of the completed models follows the same ordering (Fig.~\ref{fig:benchmark}b). We find that the mean absolute deviation of the CG density from its AA reference grows from $1.2 \pm 0.7\%$ for the hydrocarbons to $3.8 \pm 2.4\%$, $4.3 \pm 1.1\%$, and $5.8 \pm 2.0\%$ for the heteroatom, side-chain, and multifunctional levels, and is $4.0 \pm 1.4\%$ for the copolymers. These errors are averaged over all tasks that ran to completion rather than over passing tasks alone. We therefore read the two metrics jointly at matched difficulty, since a conditional accuracy computed over completed tasks alone would be inflated by the selective completion of the easier chemistries.

The resulting models deliver the efficiency for which coarse-graining is chosen in the first place. As shown in Fig.~\ref{fig:benchmark}c, CG equilibration completes in approximately 1~min of wall-clock time at every level, whereas the corresponding AA reference simulations require $38.0 \pm 8.5$ to $88.0 \pm 24.7$~min at matched system size and simulated time, which corresponds to a speed-up of roughly forty- to ninety-fold and is consistent with the fifty-fold gain observed for the PAN session. Importantly, the AA time itself rises from Level 1 to Level 4, since a more complex repeat unit contains more atoms and, once heteroatoms are present, requires explicit partial charges and long-range electrostatics. The chemistries that are hardest to type correctly are therefore also the ones whose atomistic simulation is most expensive, so coarse-graining returns the largest saving precisely where the automation is most difficult. The Level 5 copolymers fall below this trend because their comonomers are drawn from the simpler chemistries of the lower levels.

The cost of that automation is modest (Fig.~\ref{fig:benchmark}d). The LLM overhead per task ranges from $25.0 \pm 0.0$ to $32.3 \pm 0.9$ calls at a monetary cost of $(2.0 \pm 0.2)\times10^{-3}$ to $(7.0 \pm 0.7)\times10^{-3}$~\$, which is negligible relative to the compute cost of the simulations themselves. The copolymer level is the most expensive because each comonomer passes through the self-correction loop independently, yet even this amounts to seconds of LLM reasoning against tens of minutes of simulation. Taken together, these results demonstrate that the expert-dependent work of deriving a faithful CG model is performed automatically at a reasoning cost that is small compared with the simulation it enables.

\subsection{Reproducibility and robustnes of CGMas}
\label{subsec:robust}

The results reported so far were obtained under a single set of construction parameters, with one reasoning backbone. We therefore asked whether these results reflect a property of the framework or of that particular configuration. Three sources of variation are relevant: the reasoning layer is not strictly deterministic even at zero temperature, the physical conditions of the reference simulation are user-specified, and the reasoning backbone is interchangeable. We repeated a representative set of tasks under identical settings, varied the construction and simulation parameters for a fixed chemistry, and substituted the backbone, in each case running five repetitions.

We first repeated one representative task from each difficulty level five times under identical settings (\Cref{tab:reproducibility}). All five runs passed at every level, and the deviation of the CG density from its all-atom reference varied by at most $0.03$ percentage points across repetitions, from $1.70 \pm 0.012\%$ for polyacetylene to $4.40 \pm 0.002\%$ for polycaprolactam. Notably, the cost is far less reproducible than the density, ranging from ($1.86 \pm 0.03$) to ($3.78 \pm 0.97) \times 10^{-3}$~\$, and the relative spread grows with chemical complexity, reaching 26\% for polycaprolactam. Since the construction seed is held fixed, the simulation is deterministic for a given topology, so the residual variation in density reflects small differences in the generated topology and fitted parameters rather than sampling noise. These results suggest that the run-to-run uncertainty of CGMas enters almost entirely through the reasoning layer, where it manifests as a variable number of self-correction cycles, and is largely absorbed before it reaches the physical output.

\begin{table}[H]
  \small
  \centering
  \caption{Run-to-run reproducibility of CGMas. One representative task per difficulty level was repeated five times under identical settings.}
  \label{tab:reproducibility}
  \begin{tabular}{llccc}
    \toprule
    Level & Representative task & Pass & Density error (\%) & Cost ($10^{-3}$~\$) \\
    \midrule
    L1 & Polyacetylene & \checkmark & $1.70 \pm 0.012$ & $1.86 \pm 0.03$ \\
    L2 & Polyvinyl alcohol & \checkmark & $4.12 \pm 0.008$ & $3.87 \pm 0.24$ \\
    L3 & Polyacrylonitrile & \checkmark & $3.90 \pm 0.029$ & $1.97 \pm 0.10$ \\
    L4 & Polycaprolactam & \checkmark & $4.40 \pm 0.002$ & $3.78 \pm 0.97$ \\
    L5 & SBS triblock & \checkmark & $2.05 \pm 0.005$ & $6.58 \pm 0.37$ \\
    \bottomrule
  \end{tabular}
\end{table}

We next varied the physical conditions of the reference simulation while holding the chemistry fixed, changing the equilibration temperature and the way a given number of repeat units is distributed among chains (\Cref{tab:sensitivity}). Every configuration passed. The density error remained between $1.29$ and $2.99\%$, with the largest deviation at 350~K, which may be attributed to the melt being furthest from the state point at which the potentials were derived, and the smallest deviations occurring for the larger systems of 400 repeat units. Importantly, the reasoning cost was insensitive to these changes, varying only between ($1.84$--$1.98) \times 10^{-3}$~\$ across the seven configurations, which indicates that the cost of automation is set by the chemistry of the repeat unit and not by the size or the temperature of the system it is asked to build. The variation that does appear in the density is of the magnitude expected from coarse-graining at a different state point, and is therefore attributable to the CG model rather than to the framework that generated it.

\begin{table}[H]
  \small
  \centering
  \caption{Sensitivity of CGMas to the construction and simulation parameters for polyethylene at one bead per monomer. $N_{\mathrm{m}}$ is the number of repeat units per chain, $N_{\mathrm{c}}$ the number of chains, and $T$ the equilibration temperature. Each condition varies the temperature or the chain composition from the default, with all other settings held at their default values. Each condition was repeated five times.}
  \label{tab:sensitivity}
  \begin{tabular}{lcccccc}
    \toprule
    Conditions & $N_m$ & $N_c$ & $T$ (K) & Pass & Density error (\%) & Cost ($10^{-3}$ \$) \\
    \midrule
    Default    & 10 & 20 & 300 & \checkmark & 1.80 $\pm$ 0.016 & 1.85 $\pm$ 0.03 \\
    Lower $T$  & 10 & 20 & 250 & \checkmark & 1.42 $\pm$ 0.015 & 1.92 $\pm$ 0.01 \\
    Higher $T$ & 10 & 20 & 350 & \checkmark & 2.99 $\pm$ 0.020 & 1.84 $\pm$ 0.03 \\
    Longer, fewer chains  & 20 & 10 & 300 & \checkmark & 2.01 $\pm$ 0.015 & 1.87 $\pm$ 0.02 \\
    Longer chains         & 20 & 20 & 300 & \checkmark & 1.54 $\pm$ 0.011 & 1.89 $\pm$ 0.02 \\
    Longest, fewer chains & 40 & 10 & 300 & \checkmark & 1.29 $\pm$ 0.010 & 1.98 $\pm$ 0.01 \\
    More chains           & 10 & 40 & 300 & \checkmark & 1.40 $\pm$ 0.008 & 1.87 $\pm$ 0.02 \\
    \bottomrule
  \end{tabular}
\end{table}

We then repeated the polyethylene task with five LLM backbones representing a wide range of model sizes and providers, again five times each (\Cref{tab:llm}). All five backbones completed the full pipeline and passed validation, with density errors confined to a narrow band from $1.80$ to $2.55\%$ and a comparable number of LLM calls, from $20.3 \pm 3.3$ to $27.7 \pm 2.4$. In contrast, the monetary cost varies by a factor of roughly 25, from ($1.85 \pm 0.03$) to ($46.68 \pm 2.23) \times 10^{-3}$~\$. Across all three tests the physical output is stable while the cost is not, which locates the framework's variability in the reasoning layer and its reliability in the deterministic tools and the validation loop. A practical consequence is that the least expensive backbone suffices for routine use, which we adopted as the default.

\begin{table}[H]
  \small
  \centering
  \caption{Robustness of CGMas to the LLM reasoning backbone on the accuracy and cost. The polyethylene task at one bead per monomer was repeated five times with each backbone under identical settings.}
  \label{tab:llm}
  \begin{tabular}{lccccc}
    \toprule
    & 5.4-nano & 5.4-mini & 5.5 & Haiku 4.5 & Sonnet 5 \\
    \midrule
    Pass               & \checkmark & \checkmark & \checkmark & \checkmark & \checkmark \\
    Density error (\%) & 2.13 $\pm$ 0.020 & 1.80 $\pm$ 0.016 & 2.55 $\pm$ 0.002 & 2.46 $\pm$ 0.006 & 2.49 $\pm$ 0.010 \\
    LLM calls          & 27.7 $\pm$ 2.4 & 25.0 $\pm$ 0.0 & 26.3 $\pm$ 1.8 & 21.0 $\pm$ 2.9 & 20.3 $\pm$ 3.3 \\
    Cost ($\times 10^{-3}$ \$) & 2.15 $\pm$ 0.06 & 1.85 $\pm$ 0.03 & 6.91 $\pm$ 2.53 & 28.74 $\pm$ 4.76 & 46.68 $\pm$ 2.23 \\
    \bottomrule
  \end{tabular}
\end{table}

% ─── 3. Discussion ───────────────────────────────────────────────────────────
\section{Discussion}
\label{sec:discussion}

CGMas shows that a multi-agent framework, pairing LLM-based chemical reasoning with purpose-built physics-aware tools, can carry the bottom-up coarse-graining pipeline from a bare polymer name to a validated CG simulation without a user-supplied topology or manual scripting. Deriving the CG force field is the central contribution. Because the CG resolution is a design choice rather than a fixed property of the polymer, there is no readily available parameter set to look up, and every element of the model follows from the mapping the user asks for. CGMas closes that loop in one step, translating a plain-language statement of the desired resolution into a bead definition, enumerating the interaction types it implies, accumulating the corresponding distributions from the mapped atomistic trajectory, inverting and fitting them, and screening the parameters before the CG model is assembled.

Autonomous topology generation is what removes the need for user input, since the distributions can only be accumulated from a chemically correct atomistic trajectory. Its self-correction loop returns the specific violated constraint together with its local bonding context rather than a corrected assignment, which is why we expect the same machinery to transfer across chemistries. Transfer is bounded, however, by the parameter set itself: the agent selects among the atom types supplied in its prompt and introduces no new ones, so every assignment is traceable to a tabulated force field and none is invented by the language model. Exotic functional groups, metal-containing repeat units, and charged monomers such as polyelectrolytes therefore fall outside the present OPLS-AA table.

The failures are as informative as the successes. All 27 tasks ran to completion, so the five that did not pass exceeded the 5\% density band rather than terminating. Three of them carry oxygen in the backbone or in an ester linkage and show the largest deviations, at 8.96\% for poly(propylene oxide), 8.63\% for poly(lactic acid), and 5.40\% for the periodic poly(ethylene-co-vinyl alcohol). In these repeat units the partial charges of the oxygen and its neighboring carbons balance against one another, so an assignment that satisfies charge neutrality overall can still distribute the charge incorrectly, and the resulting cohesive energy carries through the mapping into the CG density. The remaining two are the styrenics, at 5.57\% for polystyrene and 6.49\% for poly($\alpha$-methylstyrene), which exceed the band narrowly and place a phenyl ring in a bead of unusually large collision diameter. Notably, the failures cluster on polar and bulky side groups rather than on the most complex backbones, since poly(caprolactam) passed at 4.40\% despite placing more distinct environments in one repeat unit than either ester.

The three robustness tests suggest that the variability of CGMas sits in the reasoning layer while its reliability sits in the deterministic tools. The density error is reproducible to within $0.03$ percentage points across repeated runs, whereas the cost varies by up to 26\% for the same task (\Cref{tab:reproducibility}) and by a factor of roughly 25 across backbones (\Cref{tab:llm}). Since the construction seed is held fixed, a variable number of self-correction cycles appears to be absorbed before it reaches the physical output. This is also why we adopted \texttt{gpt-5.4-mini} as the default, which was the least expensive backbone we tested and gave the smallest density error.

In terms of methodological limitations, direct Boltzmann inversion reproduces the sampled distributions by construction but does not self-consistently optimize the pair correlation, so structural agreement is not guaranteed where the density is matched. We adopted it here because a single-pass inversion is deterministic and has few failure modes, which is useful in a first autonomous pipeline, whereas iterative Boltzmann inversion and force matching add convergence failure as a new mode. Extending the framework to either would be a natural next step where the pair correlation is the quantity of interest.

On the reasoning side, the chemical reasoning available to the agent can be extended beyond what a single prompt can carry. Atom types are supplied at design time, so coverage is fixed in advance, and a difficult chemistry is resolved by repeated diagnosis rather than by drawing on what is already established for that class of polymer. Retrieval-augmented generation over force-field databases and the primary literature would let coverage adapt to the chemistry at hand, while organizing that literature into a structured knowledge base would let established parameterizations inform the typing of chemically related repeat units. Both directions may extend chemical coverage and reduce correction cycles.

At a broader level, CGMas illustrates one strategy for AI-augmented scientific computation, which is to embed a flexible reasoning layer within a deterministic and physics-constrained workflow that checks correctness at every stage. The architecture can be generalized along two axes. Coarse-graining is not confined to polymers, and the same pipeline could be applied to other material systems such as ionic liquids, organic semiconductors, or thermosets. The physics being examined can also be broadened, since a validated CG model, once obtained at low cost, is a starting point for questions that are impractical at the atomistic scale, such as interfacial behavior, mechanical deformation, or phase behavior. A density criterion that suffices for a polymer melt does not carry over to an interface or to a mechanical response, so each direction would need its own basis for evaluation. Hence, assembling benchmarks whose tasks and acceptance criteria are agreed across materials classes and target properties may prove as valuable as the automation itself.

% ─── 4. Methods ──────────────────────────────────────────────────────────────
\section{Methods}
\label{sec:methods}

\subsection{Agentic Framework}
\label{subsec:methods_framework}

A task in CGMas is defined by a polymer name and an optional plain-language description of the desired bead resolution, from which the framework executes the pipeline through to the coarse-grained simulation. CGMas is constructed using the LangChain and LangGraph software framework using the LangChain and LangGraph libraries. The workflow is organized as a state machine whose nodes are of two types. Chemical reasoning, natural-language parsing, and review are delegated to the LLM, whose output is variable and requires validation. Topology assembly, melt construction, mapping, distribution accumulation, potential inversion, and simulation are performed by deterministic tools, which produce reproducible results for a fixed input. The agents are stateless, so all intermediate results are held in a single typed record that is threaded through every stage.

Three loops return the workflow to an earlier stage when a check fails, and each is bounded: topology generation allows at most five regeneration attempts, the review of the assembled all-atom data file allows at most 2 revisions, and the review of the derived CG potentials allows at most two revisions. All simulations were performed on a single workstation with an Intel Core i7-14700K CPU (20 cores, 28 threads) using LAMMPS (29 Aug 2024) on 12 MPI processes~\citep{plimpton1995lammps}.

\subsection{Topology generation and self-correction pipeline}
\label{subsec:methods_topology}

Given a polymer name, the topology agent generates a JSON description of the constitutional repeat unit (CRU). The  OPLS-AA table is included in the system prompt, listing bond types, masses, partial charges, Lennard-Jones $\varepsilon$ and $\sigma$. The table was compiled from the OPLS-AA parameter set~\citep{jorgensen1996opls} as distributed with GROMACS~\citep{berendsen1995gromacs}. The agent therefore selects among existing parameters rather than proposing values, and the assignments it produces are evaluated against this table. The agent identifies the smallest CRU whose repetition reproduces the chain, assigns each atom to the OPLS-AA type whose annotated environment matches the local context, specifies the intramolecular bonds, and designates the head and tail atoms that serve as inter-CRU connection points.

The difficulty of this step is that chemically similar atoms can require different force-field types and partial charges depending on their local bonding environment. The identity of the neighboring atoms and the bond order determine the appropriate assignment, so the agent is asked to reason from the molecular structure rather than to reproduce a per-polymer template. First-pass topologies often violate the resulting constraints, and therefore the CRU passes through a sequence of validators before construction. A failing validator returns the quantitative discrepancy together with the relevant local bonding context rather than a corrected assignment, so that the agent re-derives the repair from the structure itself. This design is intended to let the same machinery apply to chemistries that were not encountered during development. The agent is re-invoked for a targeted repair for up to five cycles.

Four validators are applied. The first checks charge neutrality. The net charge $q_{\mathrm{net}} = \sum_i q_i$ over the CRU is compared with the expected value, which is zero for a neutral repeat unit, using a tolerance of $0.01$~e. When the two differ, the feedback reports the numerical charge error together with a per-atom charge table augmented with each atom's bonded neighbors, which localizes the error to the atoms near heteroatoms or $\pi$ systems where misassignment is most likely. The second checks hydrogen types, which are environment-specific in OPLS-AA, so that an alkene hydrogen is expected on an sp$^2$ alkene carbon and an aromatic hydrogen on an aromatic carbon. Violations are reported together with the expected bonded-carbon type and the acceptable alternatives. The third checks chain-end valence. Each head and tail atom carries one bond to the adjacent CRU, which reduces its free valence for intra-CRU hydrogens by one, and the hydrogen count on a carbon is compared with $4 - N_{\mathrm{heavy}} - N_{\mathrm{inter}}$, where $N_{\mathrm{inter}}$ is one for a chain-end atom and zero otherwise. Both excess and missing hydrogens are flagged, and either is a common source of charge imbalance. The fourth is a rule-based post-processor that corrects recurring heteroatom and halogen misassignments, including hydroxyl oxygens labeled as ether oxygens, ether carbons, alkene carbon subtypes, and chlorine type keys, among others.

After the constructor writes the LAMMPS data file, the review agent inspects the file together with the CRU definition. It checks format integrity, meaning that the header counts match the data sections and the atom-style fields are correct, force-field completeness, meaning that pair, bond, angle, and dihedral coefficients are present and physically plausible for the declared types, the chemical consistency of the atom-type assignment, and the head and tail connectivity between monomers. Charge neutrality and file completeness are established upstream and are excluded from this review. If revision is required, the feedback is appended to the topology prompt and the CRU is regenerated, for up to two cycles, after which the current file is retained with a warning.

Copolymers are specified by an architecture type which is random, alternating, block, and periodic, together with monomer mole fractions, block lengths, and repeat patterns. The topology agent applies the self-correction pipeline independently to each comonomer. The construction tool then generates the chain sequences according to the chosen architecture, builds each chain by appending repeat units in sequence order, and merges the atom-type tables of all comonomers into a unified global index. The size of all-atom box is set from the fraction-weighted average monomer mass.

\subsection{All-atom construction and equilibration}
\label{subsec:methods_aa}

The all-atom constructor builds an amorphous, multi-chain LAMMPS data file from the approved CRU and the confirmed simulation parameters, which are the number of chains $N_c$ (default 20), the chain length $N_m$ (default 10 repeat units), the initial packing density $\rho$ (default 0.9~g\,cm$^{-3}$), and a random seed. Coordinates are generated chain by chain by breadth-first traversal of the intramolecular bond graph with Z-matrix placement. Bond lengths and angles are read from the OPLS-AA tables, and torsions are sampled uniformly from $[0,2\pi)$ to produce amorphous conformations. Ring systems, detected as back-edges in the traversal, are placed as regular polygons oriented by a random rotation about the ring-attachment bond, which handles benzene and other cyclic groups at arbitrary side-chain depth. For copolymers, the constructor first generates the chain sequences according to the specified architecture, then builds each chain by appending repeat units in sequence order, and merges the atom-type tables of all comonomers into a unified global index. Chains are placed sequentially into a cubic periodic box of side length
\begin{equation}
  L = \left(\frac{N_c N_m M_m}{N_A\,\rho}\right)^{1/3}
  \label{eq:box_m}
\end{equation}
where $M_m$ is the repeat-unit molar mass, taken as the fraction-weighted average over the comonomers of the copolymers, and $N_A$ is Avogadro's number. Steric overlaps are avoided by an adaptive overlap-rejection scheme. For each newly placed atom the torsion angle is resampled, up to a few hundred attempts, until the atom clears all previously placed atoms. The minimum allowed separation is set per atom type, with a smaller threshold for hydrogens than for heavier atoms, so that physically tight but acceptable contacts are retained while genuine clashes are rejected.

All simulations are performed with LAMMPS~\citep{plimpton1995lammps} under periodic boundary conditions (PBC) in all x-, y-, and z-directions. Intra- and inter-molecular interactions are described by the OPLS-AA force field~\citep{jorgensen1996opls}, with geometric mixing for both Lennard-Jones $\varepsilon$ and $\sigma$ with 15~\AA\ cutoff distance. The equations of motion are integrated with the velocity-Verlet algorithm~\citep{swope1982verlet} using a 1~fs time step.

After construction, an energy minimization with the conjugate-gradient algorithm~\citep{hestenes1952conjugate} relaxes the residual close contacts of the packed configuration before dynamics. To obtain equilibrated models, an annealing procedure was used to accelerate the equilibration of model structures. During the annealing procedure, the temperature of the initial model is gradually ramped from the operating temperature $T_{\mathrm{init}}$ (default 300~K) to the annealing peak $T_{\max}$ (default 500~K) and back over $N_{\mathrm{anneal}}$ cycles, with each ramp segment run for 100~ps by default. The system is then equilibrated in the NPT ensemble with a Nos\'e--Hoover thermostat~\citep{nose1984unified, hoover1985canonical} and barostat~\citep{martyna1994constant} at 300~K for $t_{\mathrm{equil}}$ (default 1000~ps). The room temperature for the classical MD structure is 298 K to obtain appropriate thermodynamic state~\citep{choi2026coarse}. Equilibration is assessed from the convergence of the temperature, density, and potential energy over the final portion of the NPT trajectory, and the equilibrated configuration and the NPT log are retained for mapping and density extraction.

\subsection{Bead mapping and potential derivation}
\label{subsec:methods_bead}

The user describes the desired CG resolution in natural language. The mapping agent, given the polymer name, atom count, chain length, and atom-type annotations, returns a bead definition specifying the bead types and the non-overlapping atom partitions within the repeat unit, indexed from one. A bead may span more than one repeat unit. The mapping tool then computes, for every trajectory frame, the center of mass (COM) of each bead,
\begin{equation}
  \mathbf{r}_{\mathrm{COM}} = \frac{\sum_i m_i\,\mathbf{r}_i}{\sum_i m_i}
  \label{eq:com_m}
\end{equation}
with atom positions unwrapped relative to a reference atom of the bead so that the COM is evaluated consistently across the periodic boundaries and writes the mapped CG trajectory. Next, different CG potentials are constructed from the CRU to describe the interactions among the defined beads. The total potential energy of the CG system consists of bonded and non-bonded contributions.The bonded interactions are divided into contributions from relevant conformational quantities: bond length ($l$) between two adjacent beads and bending angle ($\theta$) between three consecutive beads. The non-bonded potentials are pair types of all combinations. A deterministic post-processor adds the missing interaction terms if it is absent and completes the list of pair types, so that no interaction present in the mapped system is left without a potential.

The probability distribution functions of intramolecular degree of freedom and the radial distribution functions (RDF) of intermolecular interactions are obtained by sampling the ensemble trajectories of the sufficiently equilibrated structure to describe the conformational structure at an atomistic resolution. The bond-stretching and angle-bending potentials derived in this manner are typically given by harmonic forms, and the non-bonded interaction between beads is described using a classical empirical potential with a 12-6 Lennard-Jones (LJ) form, as the following equations:
\begin{align}
  U_{\mathrm{bond}}(l)       &= \tfrac{1}{2}\,k_r\,(l - r_0)^2
    \label{eq:hb_m} \\
  U_{\mathrm{angle}}(\theta) &= \tfrac{1}{2}\,k_\theta\,(\theta - \theta_0)^2
    \label{eq:ha_m} \\
  U_{\mathrm{LJ}}(r)         &= 4\varepsilon
    \left[\left(\frac{\sigma}{r}\right)^{12}
        - \left(\frac{\sigma}{r}\right)^{6}\right]
    \label{eq:lj_m}
\end{align}
where $k_{r}$ and $k_{\theta}$ are the bond and angle energy constant, $r_{0}$ and are $\theta_{0}$ the equilibrium bond length and bending angle, $\varepsilon$ is the potential well depth, and $\sigma$ is the equilibrium distance at which the inter-bead potential energy becomes zero.

CG potentials are obtained by Boltzmann inversion of the distributions measured from atomistic simulations. The conformational probabilities are Boltzmann inverted to derive the configurational free energy:
\begin{align}
  U_{\mathrm{bond}}(l)
    &= -k_B T \ln\!\left[\frac{P(l)}{l^{2}}\right]
    \label{eq:bi_bond} \\[2pt]
  U_{\mathrm{angle}}(\theta)
    &= -k_B T \ln\!\left[\frac{P(\theta)}{\sin\theta}\right]
    \label{eq:bi_angle} \\[2pt]
  U_{\mathrm{nb}}(r)
    &= -k_B T \ln\!\big[\,g(r)\,\big]
    \label{eq:bi_nb}
\end{align}
where  $P(l)$ and $P(\theta)$ are the bond-length and bond-angle distributions, $g(r)$ is the pair RDF, $k_{\beta}$ is the Boltzmann constant, and the $l^2$ term accounts for the degeneracy in the position of a bead located at a fixed distance $l$ from another bead. Also, $sin~\theta$ term accounts for the degeneracy in the position of a bead subtending an angle $\theta$ with the bond, and $r$ is the separation distance between the CG units. The bond stretching and bending potentials derived from such direct Boltzmann inversion of bond length and angle distributions are usually good approximations of the true underlying potentials~\citep{choi2026coarse}. The reason is that each distribution has a stiff, localized dependence on its corresponding order parameter (bond stretching or angle bending) and is therefore uncorrelated with other distributions.

The review agent then cross-checks the fitted set against the enumerated interaction types. Types present in the topology but absent from the results are added and the affected potentials are recomputed. Angle equilibria above $180^\circ$ and entries with a vanishing force constant and a vanishing equilibrium value, indicating that the type was not sampled, are removed. Bond equilibrium lengths below $1.5$~\AA\ are flagged as unphysical for beads representing several atoms, and a vanishing pair well depth is retained but reported as insufficiently sampled. The backbone-backbone-backbone angle is restored if it is missing. The interaction types specified by the user are protected from automatic removal. This review is applied for up to two cycles.

\subsection{Coarse-grained simulation and glass-transition estimation}
\label{subsec:methods_cg}

Bead masses are obtained by summing the atomic masses of the constituent atoms of each bead, and the CG box size is set from \Cref{eq:box_m} using bead-level masses, with a fraction-weighted average over the comonomers for copolymers. The CG model is populated with the derived potential above and equilibrated under the same annealing procedure as the AA system, using a 1~fs time step and a Lennard-Jones cutoff of 15~\AA.

The glass-transition temperature is obtained from a cooling trajectory performed over {100}~K to {600}~K. The annealing simulation is ran with the same potentials at a cooling rate of $10^{10}~K\,s^{-1}$. $T_g$ is extracted by fitting two straight lines to the binned $\rho(T)$ data. The scanned across the sampled range and the value minimizing the total squared residual of the two segments is retained, with the intersection of the two lines reported as $T_g$. The fit is accepted only when the high-temperature branch is steeper than the low-temperature branch, as expected for the rubbery and glassy states.

\subsection{Benchmark design}
\label{subsec:methods_benchmark}

Equilibrium densities are extracted from the NPT logs by averaging over the final third of the trajectory. The primary criterion compares the CG equilibrium density with that of the equilibrated AA reference produced in the same session. A task passes when
\begin{equation}
  \left|\frac{\rho_{\mathrm{CG}} - \rho_{\mathrm{AA}}}
             {\rho_{\mathrm{AA}}}\right| \times 100
  \leq \delta_\rho = 5\%
  \label{eq:rho_m}
\end{equation}
The glass-transition temperature is reported as a secondary quantity, again as the deviation of the CG value from the AA value obtained under the identical cooling protocol and chain length, with agreement within $\delta_{T_g} = 15$~K taken as acceptable.

The benchmark comprises 27 tasks organized as a five-level difficulty ramp (\Cref{fig:tiers}). Levels 1 and 2 use one bead per repeat unit and cover hydrocarbons (4 tasks) and heteroatom-substituted repeat units (7 tasks). Levels 3 to 5 use two or more beads per repeat unit and cover complex side chains (9 tasks), multifunctional repeat units (3 tasks), and copolymers (4 tasks). Every task was run with the default construction and equilibration parameters described above.

Three metrics are recorded. The pass rate is the fraction of tasks that complete the pipeline and satisfy \Cref{eq:rho_m} within the iteration limits. The accuracy is the density error of the completed tasks, which is reported jointly with the pass rate. The cost is the per-task simulation time together with the number of LLM calls, the token count, and the monetary cost.

% ─── Data availability ────────────────────────────────────────────────────────
\section*{Data availability}
The datasets generated and/or analyzed during the current study are not publicly available due to the large volume of raw computational data, but are available from the corresponding author on reasonable request. The generated topologies and derived coarse-grained potential parameters are provided in the Supplementary Information.

% ─── Code availability ───────────────────────────────────────────────────────
\section*{Code availability}
The underlying code for this study is available upon reasonable request from the corresponding or last author.

% ─── References ──────────────────────────────────────────────────────────────
% npj uses standard Nature referencing style (numbered, superscript, journal
% abbreviations with full stops, bold volume). Switch to `naturemag' for the
% accepted version if the .bst is available in your TeX distribution:
% \bibliographystyle{naturemag}
\bibliographystyle{unsrtnat}
\bibliography{references}

% ─── Acknowledgements ────────────────────────────────────────────────────────
\section*{Acknowledgements}
We acknowledge financial support from the InnoCORE program of the Ministry of Science and ICT (N10260002), and from the National Research Foundation of Korea (NRF) grant funded by the Korean government (MSIT) (RS-2025-16070951).

% ─── Author information ───────────────────────────────────────────────────────
\section*{Author information}
These authors contributed equally: Joohee Choi, Junhyeong Lee.

\subsection*{Authors and Affiliations}

\noindent
\textbf{Department of Mechanical Engineering, Korea Advanced Institute of
Science and Technology (KAIST), Daejeon, Republic of Korea}\\
Joohee Choi \& Seunghwa Ryu

\medskip\noindent
\textbf{KAIST InnoCORE PRISM-AI Center, Korea Advanced Institute of Science and
Technology (KAIST), Daejeon, Republic of Korea}\\
Junhyeong Lee \& Seunghwa Ryu

\medskip\noindent
\textbf{Department of AX, Korea Advanced Institute of Science and Technology
(KAIST), Daejeon, Republic of Korea}\\
Seunghwa Ryu

\subsection*{Contributions}

J.C. contributed to methodology, software, investigation, formal analysis, data
curation, visualization, and writing the original draft. J.L. contributed to methodology, software, investigation, and formal analysis. S.R. supervised the study, provided conceptual guidance, and reviewed the manuscript. All authors reviewed and approved the final manuscript.

\subsection*{Corresponding author}

Correspondence to Seunghwa Ryu.

% ─── Ethics declarations ───────────────────────────────────────────────────────
\section*{Ethics declarations}
\subsection*{Competing interests}
The authors declare no competing interests.

\end{document}